%% file: attribute_hl.tex
\documentclass[final]{cvpr}

\usepackage{times}
\usepackage{epsfig}
\usepackage{graphicx}
\usepackage{amsmath}
\usepackage{amssymb}
\usepackage{color, colortbl}
\usepackage[belowskip=10pt,aboveskip=10pt]{subcaption}
\usepackage{placeins}

\newcommand{\nm}{\textcolor{black}}
 
\definecolor{Gray}{gray}{0.9}

\usepackage[pagebackref=true,breaklinks=true,colorlinks,bookmarks=false]{hyperref}

\def\cvprPaperID{49} 
\def\confYear{CVPR 2021}

\begin{document}

\title{Scalable Visual Attribute Extraction through Hidden Layers of a Residual ConvNet}

\author{Andres Baloian\\
DCC-University of Chile\\
Av. Beauchef 851, Santiago, Chile\\
{\tt\small abaloian@dcc.uchile.cl}
\and
Nils Murrugarra-Llerena\\
Snap Research\\
2772 D. Douglas L. N, S. Monica, CA, US\\
{\tt\small nmurrugarraller@snap.com}
\and
Jose M. Saavedra\\
Impresee Inc.\\
600 California, San Francisco, US\\
{\tt\small jose.saavedra@impresee.com}

}

\maketitle

\begin{abstract}
Visual attributes play an essential role in real applications based on image retrieval. For instance, the extraction of attributes from images allows an eCommerce search engine to produce retrieval results with higher precision. The traditional manner to build an attribute extractor is by training a convnet-based classifier with a fixed number of classes. However, this approach does not scale for real applications where the number of attributes changes frequently. Therefore in this work, we propose an approach for extracting visual attributes from images, leveraging the learned capability of the hidden layers of a general convolutional network to discriminate among different visual features. We run experiments with a resnet-50 trained on Imagenet,  on which we evaluate the output of its different blocks to discriminate between colors and textures. Our results show that the second block of the resnet is appropriate for discriminating colors, while the fourth block can be used for textures. In both cases, the achieved accuracy of attribute classification is superior to $93\%$. We also show that the proposed embeddings form local structures in the underlying feature space, which makes it possible to apply reduction techniques like UMAP, maintaining high accuracy and widely reducing the size of the feature space.
\end{abstract}

\input{intro}
\input{related}
\input{approach}
\input{results}
\input{conclusion}

{
    \small
    \bibliographystyle{ieee_fullname}
    \bibliography{egbib}
}

\end{document}

%% file: intro.tex
\section{Introduction}
 \label{sec:intro}
Advances in deep learning have made it possible to bring innovative solutions to the \nm{industry}
, particularly with solutions that involve computer vision. For instance, content-based image retrieval is one of many tasks that these advances have benefited.  Besides, image retrieval is a critical component in modern search engines, which allows us to retrieve relevant images given a user's query. This query can come in the form of text or image or as a combination of them.  In this vein, eCommerce is a field that leverages this kind of technology. Indeed, an effective search engine for retrieving products fitting a user's query is the first step of a successful user's journey in an online store. Thus, a practical, user-friendly, and enjoyable search engine is being increasingly demanded. 

Searching by text is the primary querying modality used by eCommerce search engines. However, its effectiveness depends on a detailed description of the product in a store. Also, to make the querying process easier, some engines include the alternative of searching by images \cite{dubey2020decade}. A special kind of this modality is when users express their intention through a drawing that leads the sketch-based image retrieval task \cite{Saavedra:2015, article:Bui_2018}. In fact, sketches represent a natural way of communication between human beings, and mobile devices' massification powers its use. 

Visual attributes play an essential role in eCommerce, as they represent specific information of what users are needing. Unfortunately, many of the products we find in an eCommerce are not described enough, lacking information as color and texture, among others. Consequently, the accuracy of the retrieval results decreases. Therefore,  including a visual attribute extractor within the search engine increases the chance the products fit the user's query. Visual attribute extraction benefits not only the searching by text modality but also the searching by images. Extracting attributes from image queries can produce more accurate results as attributes provide a high-level of semantic of a user's desire.

We can build a visual attribute extractor by collecting images representing the attribute set we are interested in extracting and then training a model with the collected data.  The problem with this simple approach is that it requires a large set of images, and it does not scale to a dynamic set of attributes. If we need to incorporate new attributes, we will need to retrain the model with the new data, which is not feasible in real\nm{-world} applications.

Therefore, the contribution of this work is to propose a scalable method for extracting visual attributes from images based on convolutional networks. Instead of training a specific network to fit a fixed number of attributes, we leverage the capability of a  convolutional network to learn visual information through its hidden neurons when it is trained in a large dataset like Imagenet \cite{imagenet}. To this end, we use a ResNet-50 \cite{he_deep_cvpr_2016} as the general convnet.

As we a priori do not know how different layers behave in front of different attributes, we conduct different experiments to determine the hidden layer's capability to classify visual information.  We present a study for two types of visual attributes, color and textures. 

To produce a trainless classifier, we take advantage of the kNN classifier. In this way, we will need to collect few image examples associated with each attribute we are interested in classifying. Then, the attribute class of a new image is inferred by the representative class of its nearest neighbors, so it is important that the space generated is characterized by forming local structures grouping elements of the same class.

Given the mentioned above, we can additionally reduce the dimension of the underlying feature space through a local-topology preserving reduction technique like UMAP \cite{leland_umap_2018}). This allows us to produce high accuracy in a low-dimension space. Indeed, after our experimental stage, we detect that it is possible to work in an 8-dimension without decreasing the original space's accuracy.

This document is organized as follows. Section \ref{sec:related_work} describes the related work. Section \ref{sec:proposal} is devoted to describing our approach in detail. Section \ref{sec:experiments} describes the evaluation experiments, and finally Section \ref{sec:conclusions} presents the conclusions.
 

%% file: related.tex
\section{Related work}
\label{sec:related_work}

\subsection{Attribute learning}
Attributes are naturally represented in a multi-task environment. Hence, \cite{shao2015deeply} learn attributes jointly for crowd scene understanding. \cite{Fouhey16} recognize 3D shape attributes using a multi-label and an embedding loss. Also, Hypergraphs encapsulate instances and attributes (in the form of relations) for learning \cite{huang2015learning}.

Other approaches use localization for attribute learning. \cite{liu2015deep} learn binary face attributes using a localization component and an identity classifier. Similarly, \cite{murrugarra2017} uses human gaze templates to learn better localizable attributes. \cite{xiao2015discovering} identify visual concepts comparing attributes in sequences. \cite{singh2016end} employ a Siamese neural network with localization and ranking sub-nets to improve the previous work.

Despite the success of deep learning, many authors employ neural networks for feature extraction and use these features for learning. \cite{liang2015unified} learn a feature space using object categories,
\cite{Gan_2016_CVPR} create category-invariant features that are helpful for attribute learning, and \cite{murrugarra_2018_aaai} learn attributes borrowing knowledge from other domain attributes and using an attention mechanism.

Previous approaches use some sort of supervision to learn attributes, where acquiring labels is expensive and time-consuming. Alternatively, in this article, we aim to discover attributes ``hidden'' in blocks of neural networks in an unsupervised fashion. 
\subsection{Dissecting neural networks}
After a convolutional model is trained, neurons contained in its hidden layers gain a certain degree of specialization, responding more strongly to certain types of image features. \cite{Rafegas} shows that neurons of shallower layers have a very high selectivity for colors and other basic visual properties, while neurons of deeper layers are mostly specialized in detecting more complex features related to the different classes. 

These results are further supported by visualization techniques, which allow humans to intuitively understand what kind of features are being extracted at each layer. \cite{Zhuwei} provides an overview of different approaches to examine convolutional models in a visual, more qualitative manner. In the case of the well-known Activation Maximization technique, synthesized images indicate that deeper layers focus on extracting increasingly more abstract information. 

\cite{li2016convergent} concludes that different model instances tend to learn similar basic features when trained on the same problem, even if they are initialized with unequal parameters. This behavior is called Convergent Learning and exemplifies how certain visual patterns such as colors, silhouettes, and textures are frequently learned by convolutional neural networks. In this work, we aim to continue these research directions, specifically, identifying which blocks are mostly related to color and texture.

\subsection{Feature extraction}

Even for deep learning techniques, feature extraction of fashion items displayed on catalog images has proven to be a difficult task. This is partly because of the tedious and toilsome process of correctly annotating the several features 
from a fashion item, which usually leads to poorly annotated data.

Taking this problem into consideration, \cite{Sandeep} proposes a progressive method for training convolutional models in the context of fashion items classification that is robust to poorly annotated data. The method divides the architecture into a backbone network and several branches, each one specialized in detecting a certain feature. These branches are first trained separately and then collectively, achieving better results than a classic multiclass model trained with perfectly annotated data.

The drawback of this method is the difficulty to be extended to a new set of attributes without requiring a training stage. Our proposal deals with this kind of problem through a simple kNN classifier.

%% file: approach.tex
\section{Approach}
\label{sec:proposal}

\begin{figure}[t]
\centering
	\begin{subfigure}[b]{.21\linewidth}
    \centering
    	\includegraphics[width=.99\textwidth]{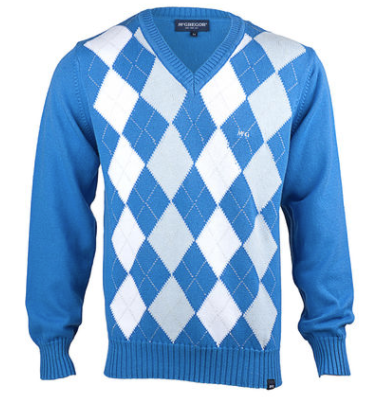} 
    \caption{Agryle}
	\end{subfigure}
	\begin{subfigure}[b]{.21\linewidth}
    \centering
    	\includegraphics[width=.99\textwidth]{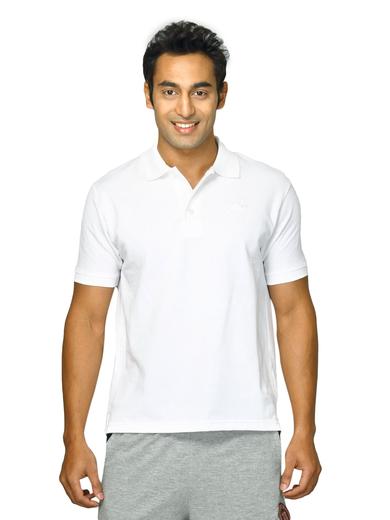}
    \caption{Basic}
	\end{subfigure}
	\begin{subfigure}[b]{.21\linewidth}
    \centering
    	\includegraphics[width=.99\textwidth]{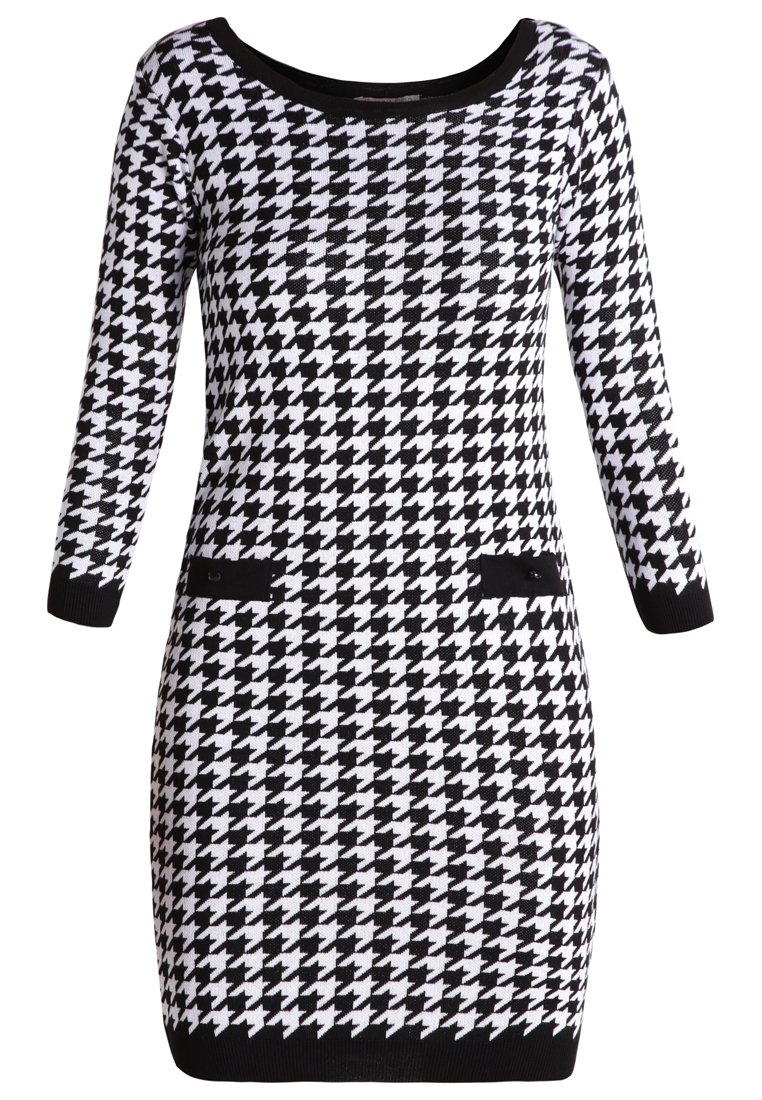}
    \caption{Crowsfeet}
	\end{subfigure}
	\begin{subfigure}[b]{.21\linewidth}
    \centering
    	\includegraphics[width=.99\textwidth]{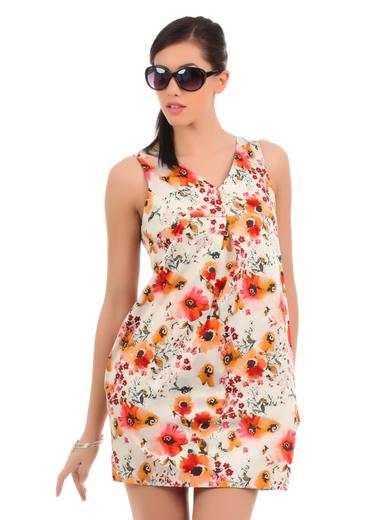}
    \caption{Flowers}
	\end{subfigure}
	\begin{subfigure}[b]{.21\linewidth}
    \centering
    	\includegraphics[width=.99\textwidth]{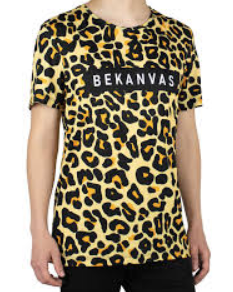}
    \caption{Leopard}
	\end{subfigure}
	\begin{subfigure}[b]{.21\linewidth}
    \centering
    	\includegraphics[width=.99\textwidth]{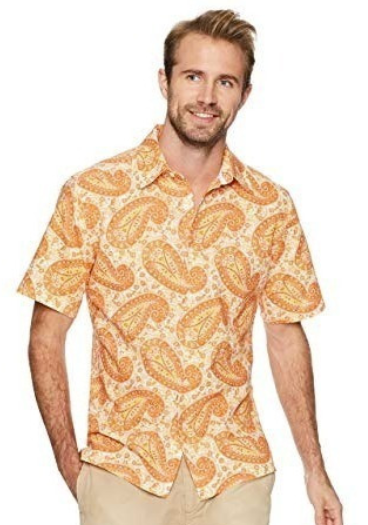}
    \caption{Paisley}
	\end{subfigure}
	\begin{subfigure}[b]{.21\linewidth}
    \centering
    	\includegraphics[width=.99\textwidth]{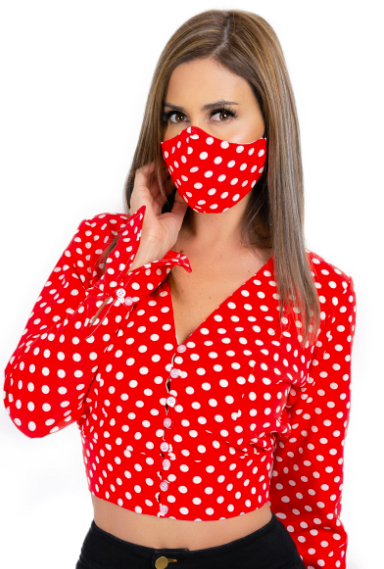}
    \caption{Polka}
	\end{subfigure}
	\begin{subfigure}[b]{.21\linewidth}
    \centering
    	\includegraphics[width=.99\textwidth]{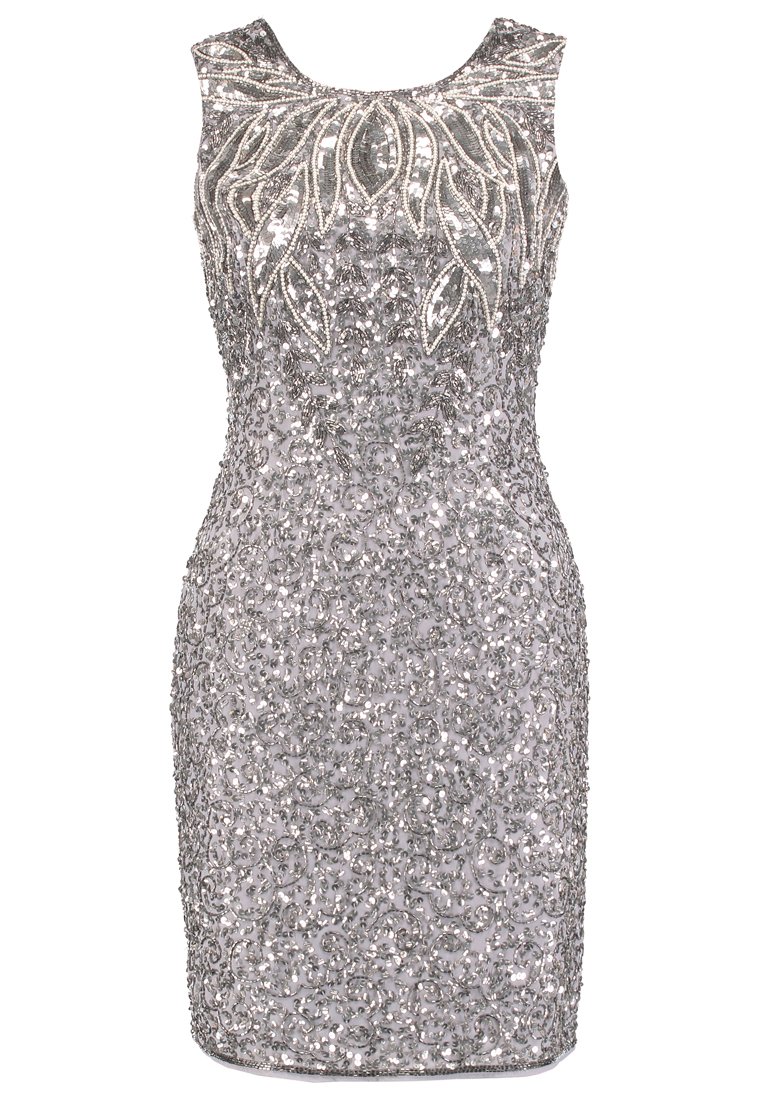}
    \caption{Sequin}
	\end{subfigure}
	\begin{subfigure}[b]{.21\linewidth}
    \centering
    	\includegraphics[width=.99\textwidth]{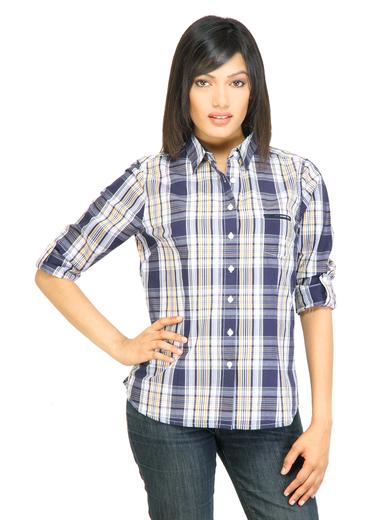}
    \caption{Squared}
	\end{subfigure}
	\begin{subfigure}[b]{.21\linewidth}
    \centering
    	\includegraphics[width=.99\textwidth]{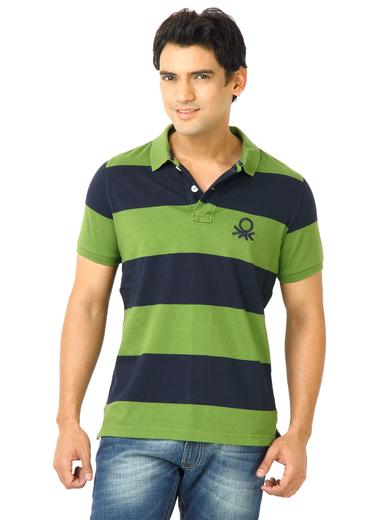}
    \caption{Striped}
	\end{subfigure}
\caption{ Ten different textures used in our dataset. }
\label{fig:example_textures}
\end{figure}

\subsection{Dataset}
\begin{figure}[t]
\centering
	\begin{subfigure}[b]{.15\linewidth}
    \centering
    \includegraphics[width=.99\textwidth]{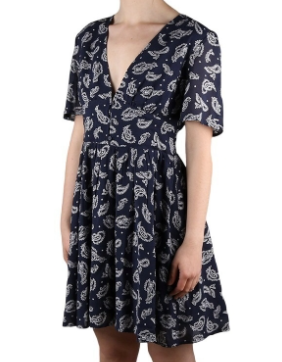} 
    \caption{Dress}
	\end{subfigure}
	\begin{subfigure}[b]{.15\linewidth}
    \centering
    \includegraphics[width=.99\textwidth]{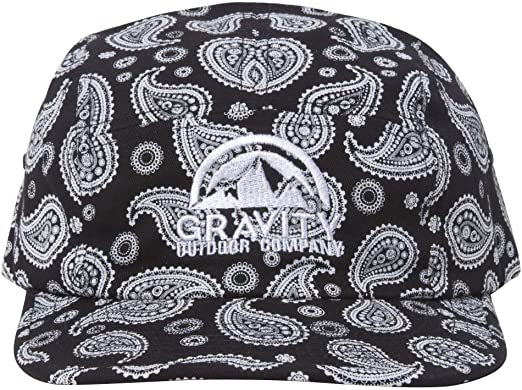}
    \caption{Hat}
	\end{subfigure}
	\begin{subfigure}[b]{.15\linewidth}
    \centering
    \includegraphics[width=.99\textwidth]{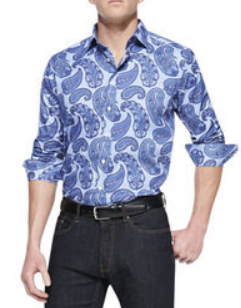}
    \caption{Shirt}
	\end{subfigure}
	\begin{subfigure}[b]{.15\linewidth}
    \centering
    \includegraphics[width=.99\textwidth]{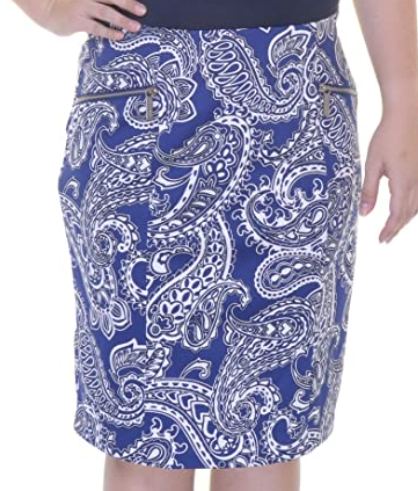}
    \caption{Skirt}
	\end{subfigure}
	\begin{subfigure}[b]{.15\linewidth}
    \centering
    \includegraphics[width=.99\textwidth]{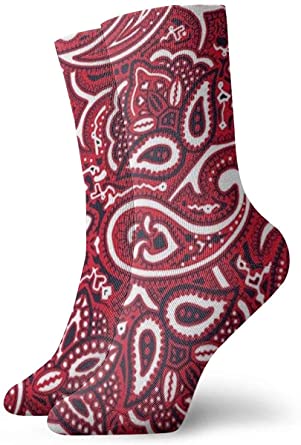}
    \caption{Socks}
	\end{subfigure}
	\begin{subfigure}[b]{.15\linewidth}
    \centering
    \includegraphics[width=.99\textwidth]{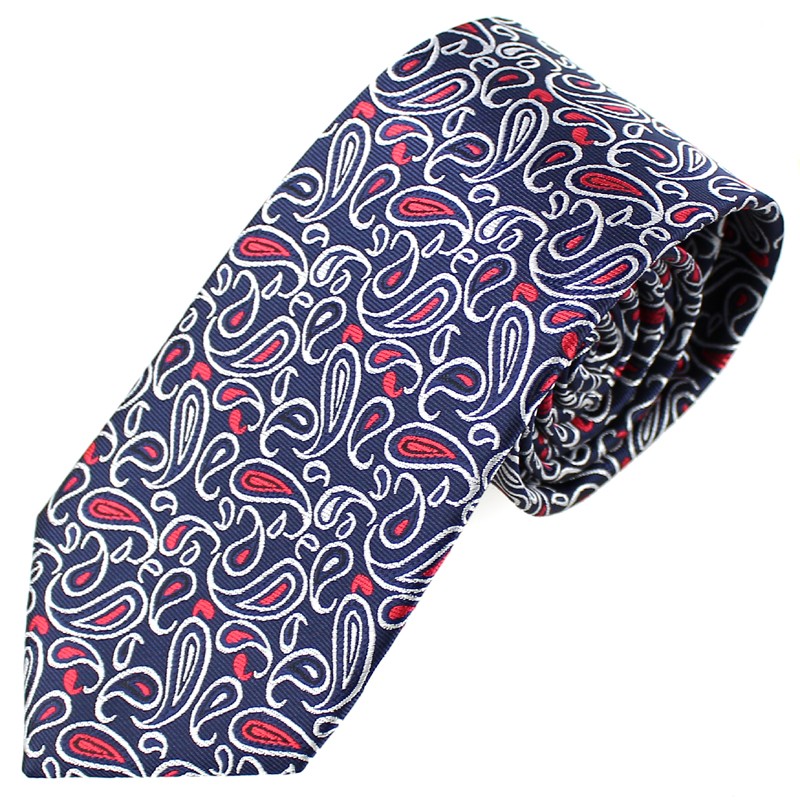}
    \caption{Tie}
	\end{subfigure}
\caption{ Sample of clothes for paisley texture. }
\label{fig:samples_paisley}
\end{figure}

We collected two datasets \nm{using Kaggle\footnote{https://www.kaggle.com/paramaggarwal/fashion-product-images-dataset} and online resources}, one grouped by color and the other by texture. The description of these datasets appears below:

\begin{itemize}
    \item \textbf{Color dataset}: We gather 100 clothing images from each of the following colors: red, black, blue, green, yellow, gray, brown, pink, purple, and orange.
    \item \textbf{Texture dataset}: We gather 100 clothing images from each of the following texture patterns: squared, striped, flowers, leopard, polka, basic, paisley, argyle, crows feet, and sequin; which are depicted in Figure \ref{fig:example_textures}. Our dataset is compiled from diverse clothes ranging from socks to shirts depicted in Figure \ref{fig:samples_paisley}.
\end{itemize}

\subsection{Feature extraction}
We selected the outputs of the five blocks of a resnet-50 \cite{he_deep_cvpr_2016} trained on ImageNet, using the Keras framework \cite{tool:keras}. For the sake of clarity, we name \emph{Block 1} the first convolution layer, and \emph{Block 2} to \emph{Block 5} are the residual blocks. 

To create a compact representation of the features, we apply a global average pooling over spatial dimension at the output of each block. Hence, we generate vectors of sizes 64, 256, 512, 1024, and 2048. These compacted vectors are fed to a machine-learning algorithm in the following section. 

\subsection{Classification}
For the classification task, we choose the kNN method over the features space generated by the residual blocks' outputs. In this way, we do not need to train the model each time a new attribute is added. Instead, we only need to collect some examples of the target attribute, and the classifier will infer the class of an image as the representative class among its nearest neighbors.

%% file: results.tex
\section{Experimental Evaluation}
\label{sec:experiments}
In this section, we provide insights on how to identify network blocks with texture and color. First, we provide details about our experimental setup. Second, we use output blocks to train a classifier for color and texture and find the most accurate one. Third, we visualize images among the blocks to observe how patterns are grouped. Finally, we use this knowledge to improve clothing retrieval.

\subsection{Setup and metrics}
We evaluate our classification experiments using a stratified 5-fold cross validation with accuracy metric and a kNN classifier\footnote{\nm{k=5 is our best parameter in our preliminary experiments}}. While for clustering, we use HDBSCAN  \cite{mcinnes2017hdbscan} on the whole dataset with euclidean distance, $min\_cluster\_size$ of 50, $min\_samples$ of 10 and $leaf\_size$ of 40. We evaluate performance using adjusted rand index \cite{hubert1985comparing} and adjusted mutual information score \cite{10.5555/1756006.1953024}.

\subsection{Quantitative experiments}
We show the behavior from outputs of five different blocks of our baseline network for grouping similar colors and textures. First, We show accuracy values for a kNN classifier in Table \ref{tab:knn}. 
For color, block two is the most discriminative (higher accuracy), while, for texture, block fourth is the most relevant.

\begin{figure}[t]
\centering
	\begin{subfigure}[b]{.53\linewidth}
    \centering
    	\includegraphics[width=.99\textwidth]{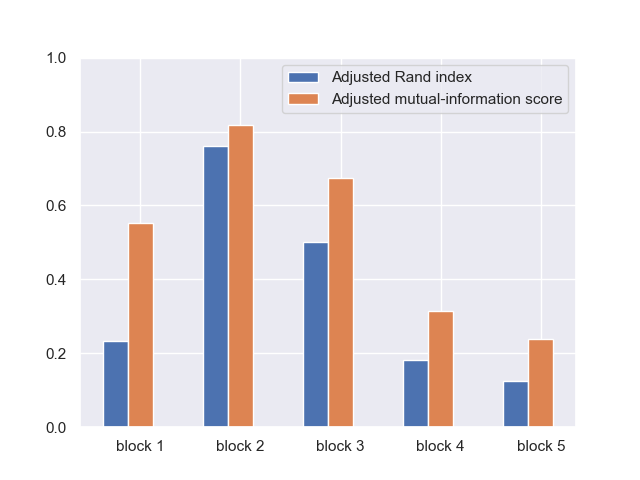} 
    \caption{Color}
	\end{subfigure}
    \hspace{-0.8cm}
	\begin{subfigure}[b]{.53\linewidth}
    \centering
    	\includegraphics[width=.99\textwidth]{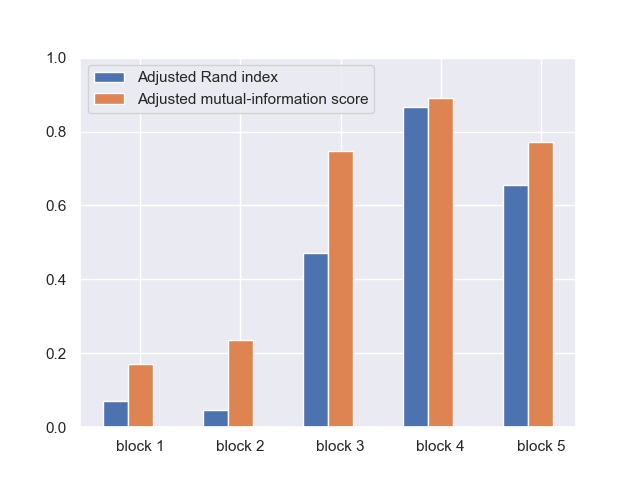}
    \caption{Texture}
	\end{subfigure}
\caption{ Cluster metrics for color and texture features among the first five blocks on a resnet-50 \cite{he_deep_cvpr_2016}.  }
\label{fig:cluster_metrics}
\end{figure}

Also, we perform a clustering study in Figure \ref{fig:cluster_metrics}. For color, we observe that that clustering metrics start increasing until block two, and then decrease consecutively in blocks three to five. On the other hand, for texture, we observe that clustering metrics increase from block one to block four.

\begin{table}[t]
    \centering
    \begin{tabular}{l|c|c}
         \textbf{\#Block} & \textbf{Color} & \textbf{Texture}  \\\hline
         Block 1& 0.900 & 0.587 \\
         Block 2& \textbf{0.937} &  0.826 \\
         Block 3& 0.929 &  0.915 \\
         Block 4& 0.886 &  \textbf{0.939}\\
         Block 5& 0.784 &  0.925
    \end{tabular}
    \caption{Accuracy achieved by the hidden layers of a resnet-50 \cite{he_deep_cvpr_2016} in the color and texture classification task using kNN with k=5}
    \label{tab:knn}
\end{table}

These insights confirm that neural networks learn low-level concepts in initial layers (e.g. color), and intermediate concepts in upper-level ones (e.g. texture).

\begin{figure*}[h!]
\centering
	\begin{subfigure}{.45\linewidth}
    \centering
    	\includegraphics[width=.99\textwidth]{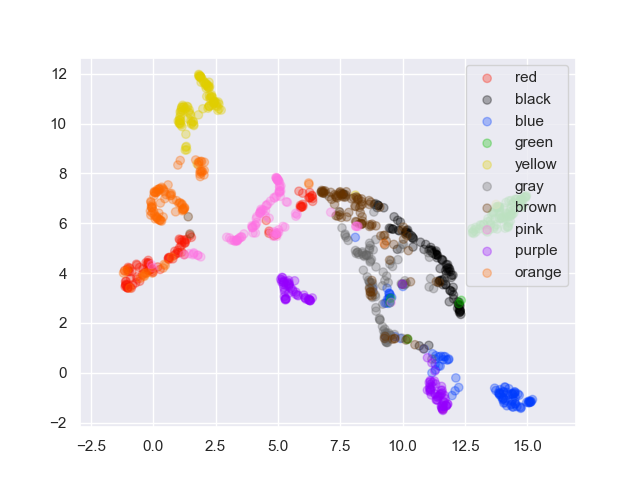} 
    \caption{Block 1}
	\end{subfigure}
    \hspace{-0.4cm}
	\begin{subfigure}{.45\linewidth}
    \centering
    	\includegraphics[width=.99\textwidth]{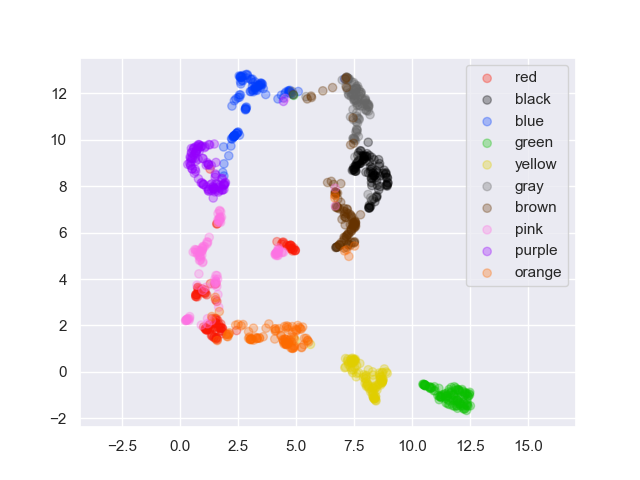}
    \caption{Block 2}
	\end{subfigure}
    \hspace{-0.4cm}
	\begin{subfigure}{.45\linewidth}
    \centering
    	\includegraphics[width=.99\textwidth]{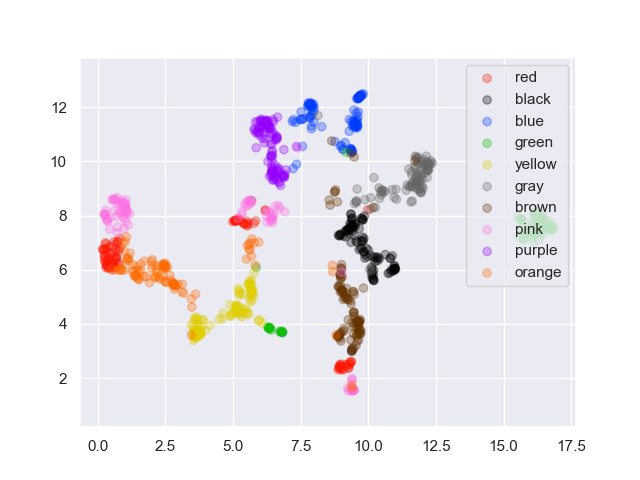} 
    \caption{Block 3}
	\end{subfigure}
	\hspace{-0.4cm}
	\begin{subfigure}{.45\linewidth}
    \centering
    	\includegraphics[width=.99\textwidth]{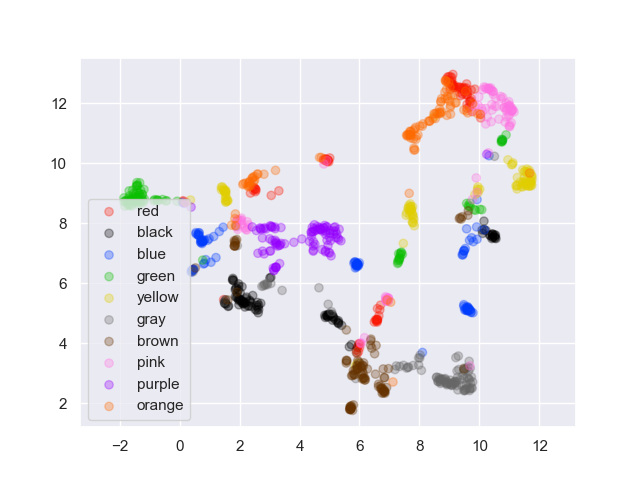} 
    \caption{Block 4}
	\end{subfigure}
	\hspace{-0.4cm}
	\begin{subfigure}{.45\linewidth}
    \centering
    	\includegraphics[width=.99\textwidth]{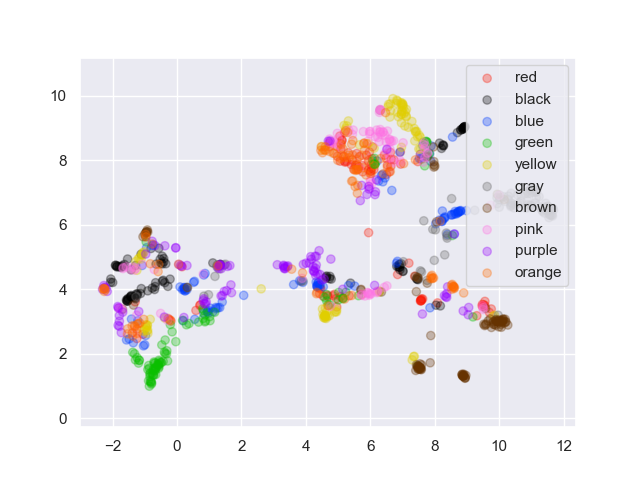} 
    \caption{Block 5}
	\end{subfigure}
\caption{UMAP visualization for color features among the first five blocks on a resnet-50 \cite{he_deep_cvpr_2016}.}
\label{fig:umap_color_blocks}
\end{figure*}


\subsection{Qualitative experiments}
\begin{figure*}[h!]
\centering
	\begin{subfigure}{.45\linewidth}
    \centering
    	\includegraphics[width=.99\textwidth]{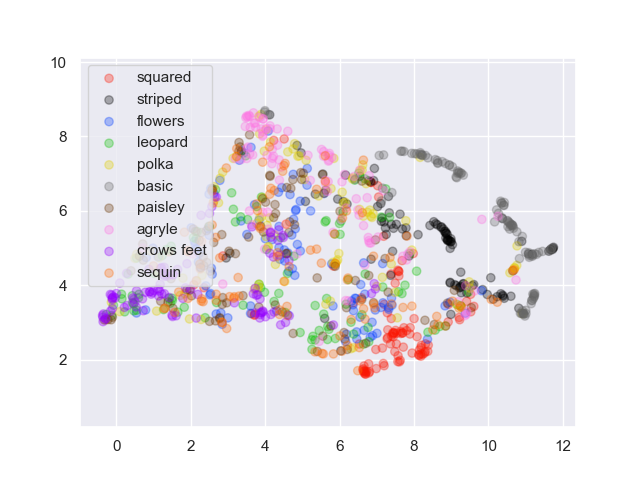} 
    \caption{Block 1}
	\end{subfigure}
    \hspace{-0.4cm}
	\begin{subfigure}{.45\linewidth}
    \centering
    	\includegraphics[width=.99\textwidth]{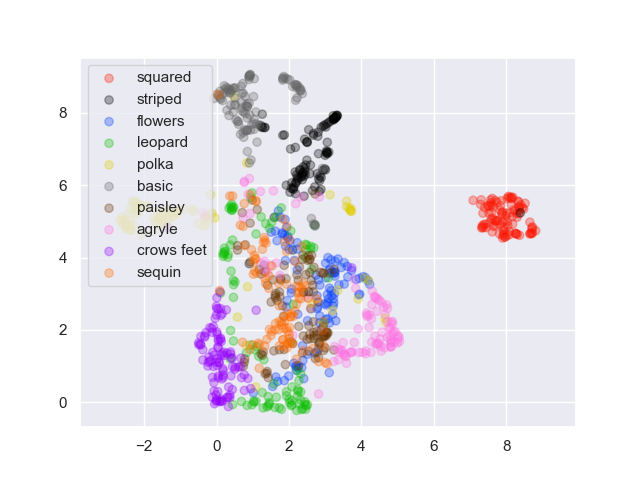}
    \caption{Block 2}
	\end{subfigure}
    \hspace{-0.4cm}
	\begin{subfigure}{.45\linewidth}
    \centering
    	\includegraphics[width=.99\textwidth]{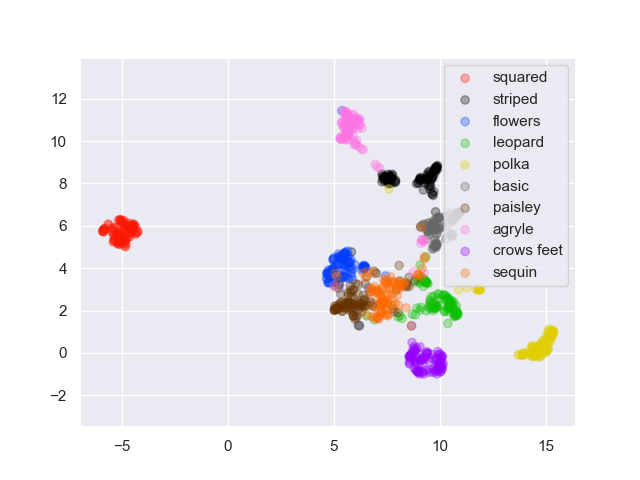} 
    \caption{Block 3}
	\end{subfigure}
	\hspace{-0.4cm}
	\begin{subfigure}{.45\linewidth}
    \centering
    	\includegraphics[width=.99\textwidth]{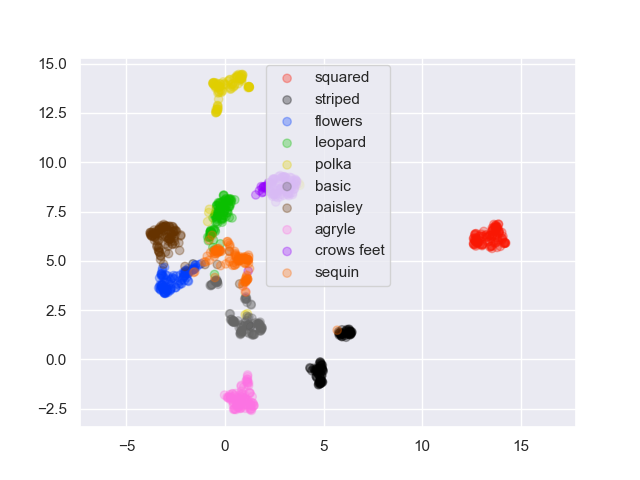} 
    \caption{Block 4}
	\end{subfigure}
	\hspace{-0.4cm}
	\begin{subfigure}{.45\linewidth}
    \centering
    	\includegraphics[width=.99\textwidth]{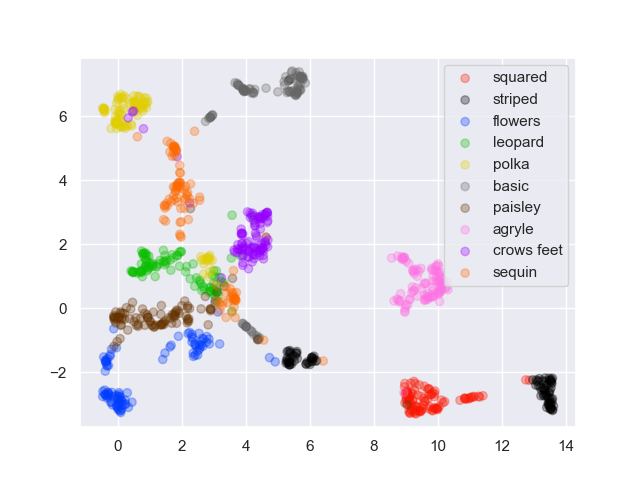} 
    \caption{Block 5}
	\end{subfigure}
\caption{UMAP visualization for texture features among the first five blocks on a resnet-50 \cite{he_deep_cvpr_2016}.}
\label{fig:umap_texture_blocks}
\end{figure*}

In order to analyze the internal behaviour of the blocks, we visualize the feature vectors using UMAP \cite{leland_umap_2018} on Figures \ref{fig:umap_color_blocks} and \ref{fig:umap_texture_blocks} for color and textures, respectively.

In Figure \ref{fig:umap_color_blocks}, we observe that block two has a clear separation among color groups, while block five spreads color features among the space and no clear groups can be found. Block two shows that yellow and green colors form their own cohesive groups, and easily differentiate from other colors. Other colors are still cohesive, however, boundaries with their neighbor group colors are less clear.

In Figure \ref{fig:umap_texture_blocks}, block four is the best to identify textures. Textures squares, stripes, argyle, and polka are the best cohesive and clear groups; while other textures still are grouped, but the boundaries with their neighbors are near. In this scenario, the worst block is block one, which shows a lot of texture overlap and no clear groups confirming that more complex patterns (i.e. textures) are learned in intermediate layers, as opposed to color, which is learned in initial layers.

\subsection{Reduced Feature Space}
Taking advantage of the local-structure generated by Block 2, for color description, and block 4, for texture, we reduce the original feature space dimensions with UMAP\cite{mcinnes2017hdbscan}, an unsupervised local-topology preserving reduction dimension technique. Table \ref{tab:umap_components} shows the classification accuracy for color and texture with different dimensions after applying UMAP. A space with 8 dimensions presents a good trade-off between accuracy and space reduction. In this case, the accuracy for color classification is 0.901 and for texture is 0.910.

\begin{table}[t]
    \centering
    \begin{tabular}{l|c|c}
         \textbf{Dimension} & \textbf{Color} & \textbf{Texture}  \\\hline
         2& 0.650 & 0.676 \\
         4& 0.892 &  0.903 \\
         \rowcolor{Gray}
         8& 0.901 &  0.910 \\
         16& 0.899 &  0.913\\
         32& 0.904 &  0.912 \\
         64& 0.894 &  0.916 \\
         128& 0.897 &  0.916 \\
         256& 0.934 &  0.915 \\
         512& -- &  0.916 \\
         1024& -- &  0.939 \\
    \end{tabular}
    \caption{Accuracy achieved by the best hidden layers of a resnet-50 \cite{he_deep_cvpr_2016} for color and texture classification using different number of dimensions from UMAP \cite{leland_umap_2018}. A dimension equal to 8 preserves similar information to the original feature space.}
    \label{tab:umap_components}
\end{table}


\subsection{Application: image retrieval}

Clothing retrieval systems usually focus on high-level semantic information, however, users like to preserve low-level features. Our findings are integrated to preserve color and texture attributes in an isolated and combined manner. 

\begin{figure*}[ht!]
\centering
	\begin{subfigure}[b]{.33\linewidth}
    \centering
    	\includegraphics[width=.99\textwidth]{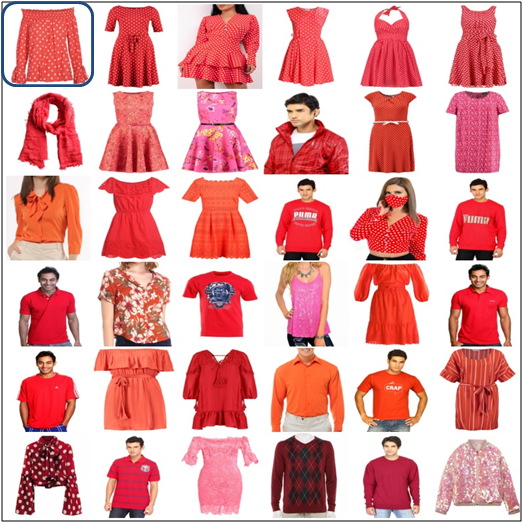} 
    \caption{Color}
	\end{subfigure}
	\begin{subfigure}[b]{.33\linewidth}
    \centering
    	\includegraphics[width=.99\textwidth]{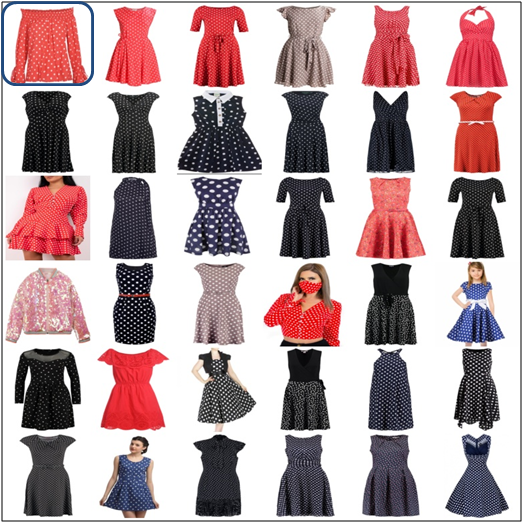}
    \caption{Texture}
	\end{subfigure}
	\begin{subfigure}[b]{.33\linewidth}
    \centering
    	\includegraphics[width=.99\textwidth]{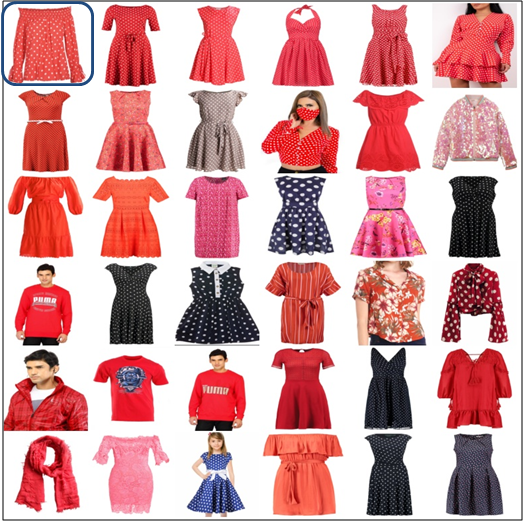}
    \caption{Color and Texture}
	\end{subfigure}
\caption{Retrieval results for a) color, b) texture and c) color and texture. The squared image on the top left corner is the query image. Retrieval results are ordered from left to right and top to bottom. Color query lacks to preserve texture starting row two. Similarly, texture query only preserves the texture and does not focus on color as some gray and blue clothes are among the top retrieved clothes. On the other hand, the last column combine color and texture, and retrieves more meaningful clothes preserving color and texture among its top results.}
\label{fig:retrieval}
\end{figure*}

We show our retrieval results in Figure \ref{fig:retrieval} for color, texture and color plus texture attributes. Our color query lacks to preserve texture starting in row two. Similarly, our texture query only preserves the texture, and does not focus on color as some gray and blue clothes are among the top retrieved clothes. On the other hand, our color plus texture query retrieves more meaningful clothes preserving color and texture among the top results. In the first row, we observe red dresses with white dots texture pattern, Also, in row two, some retrieved images are red and have white dots, or have at least one of the attributes.

%% file: conclusion.tex
\section{Conclusion}
\label{sec:conclusions}
We described an approach to identify semantic attributes among different layers in a residual neural network. Specifically, we find that block two and block four of a resnet-50 \cite{he_deep_cvpr_2016} focused on color and texture patterns. We use these findings to improve image retrieval, Importantly, by combining information from these two feature vectors, we update the ranking of retrieved results prioritizing samples with color and texture similarities.

In the future, we will investigate additional types of attributes such as parts, behavior, shape, and others. 

%% file: attribute_hl.bbl
\begin{thebibliography}{10}\itemsep=-1pt

\bibitem{Sandeep}
Sandeep~Singh Adhikari, Sukhneer Singh, Anoop Rajagopal, and Aruna Rajan.
\newblock Progressive fashion attribute extraction.
\newblock 2019.

\bibitem{article:Bui_2018}
Tu Bui, Leonardo Ribeiro, Moacir Ponti, and John Collomosse.
\newblock Sketching out the details: Sketch-based image retrieval using
  convolutional neural networks with multi-stage regression.
\newblock {\em Computers \& Graphics}, 71, 2018.

\bibitem{tool:keras}
Fran\c{c}ois Chollet.
\newblock Keras, 2015.

\bibitem{imagenet}
Jia Deng, Wei Dong, Richard Socher, Li-Jia Li, Kai Li, and Li Fei-Fei.
\newblock Imagenet: A large-scale hierarchical image database.
\newblock In {\em 2009 IEEE conference on computer vision and pattern
  recognition}, pages 248--255, 2009.

\bibitem{dubey2020decade}
Shiv~Ram Dubey.
\newblock A decade survey of content based image retrieval using deep learning,
  2020.

\bibitem{Fouhey16}
David~F. Fouhey, Abhinav Gupta, and Andrew Zisserman.
\newblock {3D} shape attributes.
\newblock In {\em Computer Vision and Pattern Recognition (CVPR)}. IEEE, 2016.

\bibitem{Gan_2016_CVPR}
Chuang Gan, Tianbao Yang, and Boqing Gong.
\newblock Learning attributes equals multi-source domain generalization.
\newblock In {\em Computer Vision and Pattern Recognition (CVPR)}. IEEE, 2016.

\bibitem{he_deep_cvpr_2016}
K. {He}, X. {Zhang}, S. {Ren}, and J. {Sun}.
\newblock Deep residual learning for image recognition.
\newblock In {\em 2016 IEEE Conference on Computer Vision and Pattern
  Recognition (CVPR)}, pages 770--778, 2016.

\bibitem{huang2015learning}
Sheng Huang, Mohamed Elhoseiny, Ahmed Elgammal, and Dan Yang.
\newblock Learning hypergraph-regularized attribute predictors.
\newblock In {\em Computer Vision and Pattern Recognition (CVPR)}. IEEE, 2015.

\bibitem{hubert1985comparing}
L. Hubert and P. Arabie.
\newblock {Comparing partitions}.
\newblock {\em Journal of classification}, 2(1):193--218, 1985.

\bibitem{li2016convergent}
Yixuan Li, Jason Yosinski, Jeff Clune, Hod Lipson, and John Hopcroft.
\newblock Convergent learning: Do different neural networks learn the same
  representations?, 2016.

\bibitem{liang2015unified}
Kongming Liang, Hong Chang, Shiguang Shan, and Xilin Chen.
\newblock A unified multiplicative framework for attribute learning.
\newblock In {\em International Conference on Computer Vision (ICCV)}. IEEE,
  2015.

\bibitem{liu2015deep}
Ziwei Liu, Ping Luo, Xiaogang Wang, and Xiaoou Tang.
\newblock Deep learning face attributes in the wild.
\newblock In {\em Computer Vision and Pattern Recognition (CVPR)}. IEEE, 2015.

\bibitem{mcinnes2017hdbscan}
Leland McInnes, John Healy, and Steve Astels.
\newblock hdbscan: Hierarchical density based clustering.
\newblock {\em The Journal of Open Source Software}, 2(11):205, 2017.

\bibitem{leland_umap_2018}
Leland McInnes, John Healy, Nathaniel Saul, and Lukas Grossberger.
\newblock Umap: Uniform manifold approximation and projection.
\newblock {\em The Journal of Open Source Software}, 3(29):861, 2018.

\bibitem{murrugarra2017}
Nils Murrugarra-Llerena and Adriana Kovashka.
\newblock Learning attributes from human gaze.
\newblock In {\em Winter Conference of Computer Vision (WACV)}, 2017.

\bibitem{murrugarra_2018_aaai}
Nils Murrugarra{-}Llerena and Adriana Kovashka.
\newblock Asking friendly strangers: Non-semantic attribute transfer.
\newblock In {\em Thirty-Second {AAAI} Conference on Artificial Intelligence,
  {AAAI} 2018}, 2018.

\bibitem{Zhuwei}
Zhuwei Qin.
\newblock How convolutional neural networks see the world - a survey of
  convolutional neural network visualization methods.
\newblock 2018.

\bibitem{Rafegas}
Ivet Rafegas, Maria Vanrella, Luis~A. Alexandre, and Guillem Arias.
\newblock Understanding trained cnns by indexing neuron selectivity.
\newblock 2019.

\bibitem{Saavedra:2015}
Jose~M. Saavedra and Juan~Manuel Barrios.
\newblock Sketch based image retrieval using learned keyshapes {(LKS)}.
\newblock In {\em Proceedings of the British Machine Vision Conference 2015,
  {BMVC} 2015, Swansea, UK, September 7-10, 2015}, pages 164.1--164.11, 2015.

\bibitem{shao2015deeply}
Jing Shao, Kai Kang, Chen~Change Loy, and Xiaogang Wang.
\newblock Deeply learned attributes for crowded scene understanding.
\newblock In {\em Computer Vision and Pattern Recognition (CVPR)}. IEEE, 2015.

\bibitem{singh2016end}
Krishna~Kumar Singh and Yong~Jae Lee.
\newblock End-to-end localization and ranking for relative attributes.
\newblock In {\em European Conference on Computer Vision (ECCV)}. Springer,
  2016.

\bibitem{10.5555/1756006.1953024}
Nguyen~Xuan Vinh, Julien Epps, and James Bailey.
\newblock Information theoretic measures for clusterings comparison: Variants,
  properties, normalization and correction for chance.
\newblock {\em J. Mach. Learn. Res.}, 11:2837–2854, Dec. 2010.

\bibitem{xiao2015discovering}
Fanyi Xiao and Yong Jae~Lee.
\newblock Discovering the spatial extent of relative attributes.
\newblock In {\em International Conference on Computer Vision (ICCV)}. IEEE,
  2015.

\end{thebibliography}
